\documentclass[twoside,leqno,twocolumn]{article}

\usepackage[letterpaper]{geometry}

\usepackage{ltexpprt}
\usepackage{hyperref}

\usepackage[utf8]{inputenc}

\usepackage{microtype}

\usepackage{amsmath}

\usepackage{graphicx}

\usepackage{booktabs}

\usepackage{enumitem}

\usepackage{multirow}

\usepackage{balance}

\begin{document}

\newcommand\relatedversion{}

\title{\Large \textsc{Gotta}: Generative Few-shot Question Answering\\ by Prompt-based Cloze Data Augmentation}
\author{Xiusi Chen\thanks{University of California, Los Angeles. xchen@cs.ucla.edu}
\and Yu Zhang\thanks{University of Illinois at Urbana-Champaign. yuz9@illinois.edu}
\and Jinliang Deng\thanks{Univ of Technology Sydney. jinliang.deng@student.uts.edu.au}
\and Jyun-Yu Jiang\thanks{Amazon Search. jyunyu@amazon.com}
\and Wei Wang\thanks{University of California, Los Angeles. weiwang@cs.ucla.edu}}

\date{}

\maketitle


\fancyfoot[R]{\scriptsize{Copyright \textcopyright\ 2023 by SIAM\\
Unauthorized reproduction of this article is prohibited}}





\begin{abstract}
Few-shot question answering (QA) aims at precisely discovering answers to a set of questions from context passages while only a few training samples are available.
Although existing studies have made some progress and can usually achieve proper results, they suffer from understanding deep semantics for reasoning out the questions. In this paper, we develop \textsc{Gotta}, a \textbf{G}enerative pr\textbf{O}mp\textbf{T}-based da\textbf{T}a \textbf{A}ugmentation framework to mitigate the challenge above. Inspired by the human reasoning process, we propose to integrate the cloze task to enhance few-shot QA learning. Following the recent success of prompt-tuning, we present the cloze task in the same format as the main QA task, allowing the model to learn both tasks seamlessly together to fully take advantage of the power of prompt-tuning. Extensive experiments on widely used benchmarks demonstrate that \textsc{Gotta} consistently outperforms competitive baselines, validating the effectiveness of our proposed prompt-tuning-based cloze task, which not only fine-tunes language models but also learns to guide reasoning in QA tasks. Further analysis shows that the prompt-based loss incorporates the auxiliary task better than the multi-task loss, highlighting the strength of prompt-tuning on the few-shot QA task.

\end{abstract}

\section*{Keywords}
question answering, knowledge base, entity, data augmentation

\section{Introduction}
\label{sec:intro}
Question answering~(QA) is the task of precisely discovering answers to natural language questions given the narrative contexts.
With a wide range of downstream applications, such as knowledge graph completion~\cite{liu2022joint}, response recommendation~\cite{zhao2013topic}, review opinion mining~\cite{zhao2019riker}, and product attribute extraction~\cite{wang2020learning}, it has drawn a lot of attention in the text mining community and has risen to one of the holy-grail tasks. Following the line of supervised learning, one can successfully build QA methods that achieve decent results. However, the assumption that a large amount of annotated QA training examples quickly poses limitations since annotation requiring efforts from domain experts is extremely expensive.

\begin{figure}[t!]
    \centering
    \includegraphics[width=.99\linewidth]{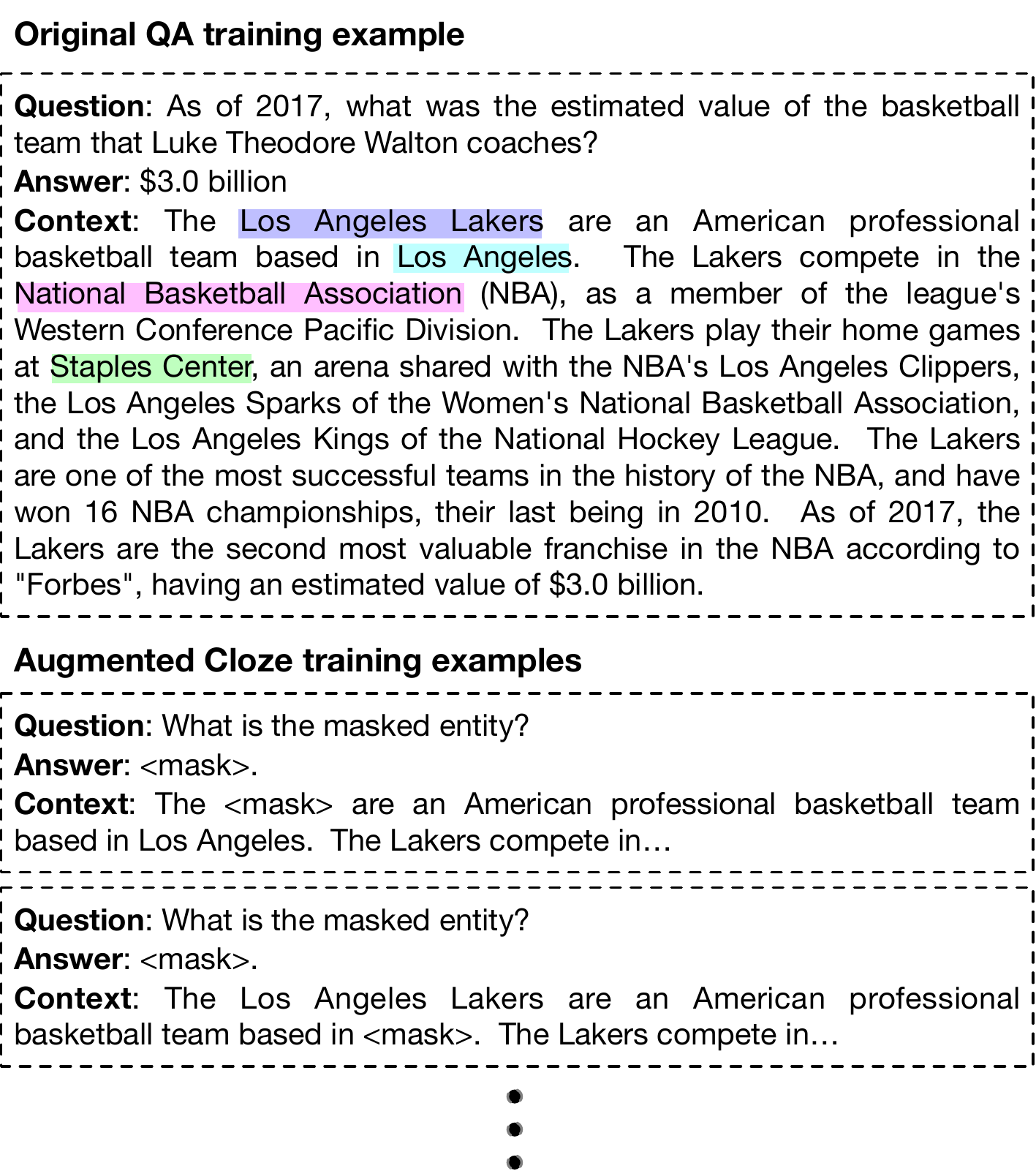}
    \caption{An example of how entity-aware text masking and prompt-style data augmentation work. \textsc{Gotta} selects entities that are covered by knowledge bases, and creates prompt-style augmented data for training purpose.}
    \label{fig:aug_data_example}
    \vspace{-5mm}
\end{figure}

We investigate the few-shot QA task, which aims to solve the QA task while only a few training examples are present. Under the few-shot setting, most existing approaches either propose a new task and pre-train a large language model from scratch~\cite{ram2021few}, or fine-tune the pre-trained model on the training examples~\cite{chada2021fewshotqa}. These practices do not explicitly understand the entities in the input text (i.e., the context text and the question text) before generating the output (i.e., the answer text), which contradicts the conventional human thinking process.
For example, in reading comprehension exams, people have to fully understand and digest the context semantics before getting precise answers.
In other words, directly mapping from the text and the question to the answer lacks a deep understanding of the context.



To bridge this gap, we develop \textsc{Gotta}, a \textbf{G}enerative pr\textbf{O}mp\textbf{T}-based da\textbf{T}a \textbf{A}ugmentation framework for few-shot QA. In \textsc{Gotta}, we design a knowledge-based cloze task to serve as a companion to enhance the main QA task. To make the cloze task more dedicated for QA, we utilize publicly available knowledge bases and focus on the covered entities by only selecting the entities in the text as the object to construct cloze problems. By constructing more data for fine-tuning, we incorporate the external knowledge in the knowledge bases in the hope of introducing more inductive bias that is beneficial to the QA task. The inductive bias provides extra supervision beyond the weak supervision signals only provided in the few-shot QA training set. Intuitively, the cloze task is to imitate the human behavior of understanding the context by filling in the blanks. We conduct this entity-aware cloze because identifying the entities and understanding their relations is crucial for solving QA problems on the same chunk of text.

\begin{figure*}[t!]
    \centering
    \includegraphics[width=\linewidth]{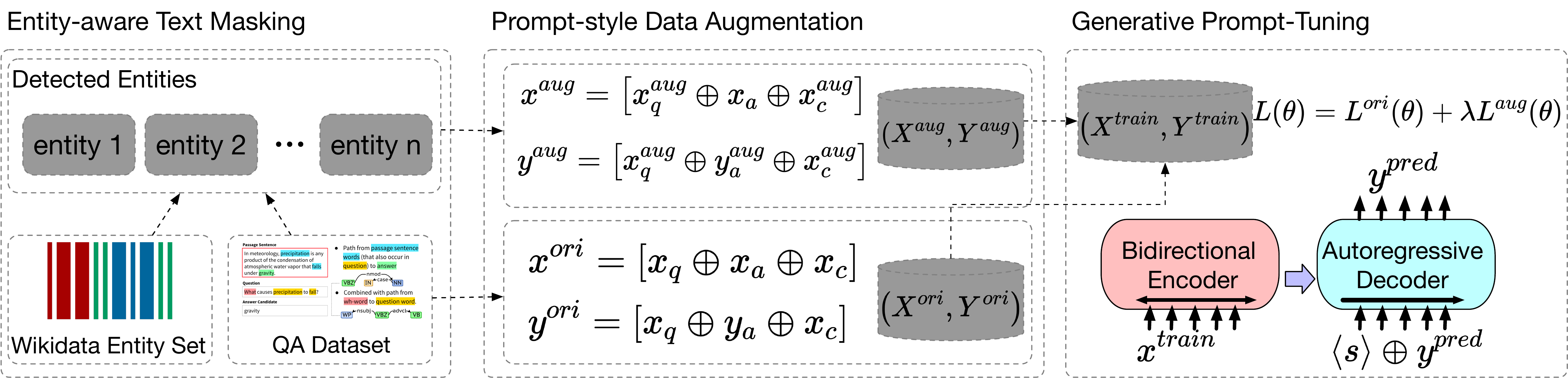}
    \caption{Framework overview for \textsc{Gotta}.}
    \label{fig:framework}
    \vspace{-2mm}
\end{figure*}

Inspired by recent advantages of prompt-tuning, as shown in Figure~\ref{fig:aug_data_example} and Figure~\ref{fig:framework}, we feature both QA and cloze tasks in the same prompt template to align with each other at the pre-training stage. Following this routine, no redundant model parameters are introduced while the pre-trained model can maximize the performance on our downstream QA task, especially under the few-shot setting.
Although our cloze task is quite similar to the popular masked language modeling (MLM), there are two major distinctions between entity-aware text masking and MLM. First, MLM randomly masks word tokens while entity-aware text masking only targets entities that are more likely to be relevant to the QA task. Second, MLM is usually pipelined with a softmax function to select one token while entity-aware text masking generates a token sequence to form a text span, which is more favored by QA tasks.
Extensive experiments on publicly available and conventional benchmarks demonstrate that \textsc{Gotta} is able to achieve generally better results over competitive baselines, validating the effectiveness of the cloze task. Further in-depth analysis shows that the prompt-based loss incorporates the auxiliary task better than classification loss, highlighting the effectiveness of prompt-tuning on the few-shot QA task.

We summarize our contributions as follows:
\begin{itemize}[leftmargin=*]
    \item We propose to incorporate the cloze task as a data augmentation module to extract self-supervised training examples to enhance the learning for few-shot QA. 
    \item We formulate both QA and cloze tasks in the same format, allowing us to apply prompt-tuning to take full advantage of pre-trained large language models.
    \item We conduct extensive experiments on publicly accessible benchmarks to validate the effectiveness of \textsc{Gotta}, and observe consistent improvement over competitive compared methods. Beyond that, we also study the necessity of different parts of the model, providing the readers with a better understanding of the framework.\footnote{The code for \textsc{Gotta} is at \url{https://github.com/xiusic/Gotta}. }
\end{itemize}

\section{Related Work}
Existing studies most related to our work come from three aspects: few-shot QA, prompt-tuning, and data augmentation. In this section, we briefly recap and distinguish our proposed method from theirs.

\vspace{1mm}

\noindent \textbf{Few-Shot QA.} Prior studies in QA either reuse the high-performing pre-trained language models (PLMs)~\cite{Lan2020ALBERT:,joshi2020spanbert}, or train a model from scratch on synthetic QA data~\cite{puri2020training,lewis2019unsupervised,alberti2019synthetic}. However, all of them require fine-tuning the models on massive annotated data from the downstream QA task, which is often impractical in real-world cases. Several approaches have recently been developed to allow the model to quickly adapt to the downstream task with solely a handful of annotated data~\cite{ram2021few,chada2021fewshotqa}. Ram et al. \cite{ram2021few} tailor the pre-training scheme specialized for handling QA tasks. They design a recurring span selection objective for pre-training, which aligns with the common objective in extractive QA tasks. To save the effort to pre-train the model on a large-scale corpus, Chada and Natarajan \cite{chada2021fewshotqa} seek to explore the capacity of the existing PLMs. They propose a simple framework, known as FewshotQA, where a QA-style prompt is constructed to cast the QA problem as a text generation problem. Specifically, the prompt is created as a concatenation of the question and a mask token representing the answer span. In this way, the input format is geared toward processing by the PLMs. Distinct from these two studies, we focus on exploring more relevant information in the context data, aside from the annotated QA pairs, to fine-tune the model under the few-shot setting. KECP~\cite{wang2022kecp} is a concurrent work with \textsc{Gotta} that focuses only on extractive QA (EQA). Also inspired by prompt-tuning, KECP views the EQA task as a non-autoregressive MLM generation problem and uses a span-level contrastive learning objective to improve the final performance.

\vspace{1mm}

\noindent \textbf{Prompt-Tuning.}
Standard fine-tuning of PLMs for few-shot learning does not achieve satisfying performance in many cases because the limited training samples may not be sufficient for optimizing the parameters in the newly introduced task head. To reuse the language modeling capability of PLMs without introducing randomly initialized parameters, prompt-based approaches~\cite{gao2021making,hu2022knowledgeable,logan2022cutting,min2022noisy,schick2021exploiting,schick2021s,tam2021improving} formulate training samples as natural language prompt templates so that the downstream tasks can be solved as a masked token prediction problem. Further studies propose to replace the manual design of prompts with automatic search or learning~\cite{cui2022prototypical,hambardzumyan2021warp,lester2021power,liu2021gpt,zhang2021differentiable,zhong2021factual}. Although prompt-tuning has demonstrated remarkable few-shot learning performance in some tasks (e.g., text classification and natural language inference~\cite{wang2019glue}), it has not been extensively explored in question answering. In this paper, we explore a prompt-based data augmentation framework for few-shot QA.

\vspace{1mm}

\noindent \textbf{Data Augmentation.}
Under the few-shot setting, data augmentation mainly aims to create more training data based on a small number of provided training samples to overcome label scarcity when training the model. Pioneering studies on text data augmentation include EDA~\cite{wei2019eda} and UDA~\cite{xie2020unsupervised}, which leverage text editing (e.g., synonym replacement, random swap) and back translation to create more labeled data for text classification. In another line of work, several studies generate training data by fine-tuning autoregressive PLMs on the training set~\cite{anaby2020not,yang2020generative} or using label-specific prompts~\cite{schick2021generating} to guide text generation toward the desired label. However, most of the studies mentioned above focus on the task of few-shot text classification. In contrast, our \textsc{Gotta} framework proposes a cloze-style data augmentation method for few-shot QA.

\vspace{-10pt}
\section{\textsc{Gotta}: The Proposed Framework}

The overall framework of \textsc{Gotta} is illustrated in Figure~\ref{fig:framework}. The core idea is to augment training data with cloze-style questions to force the model to understand the contexts beyond the original questions. We fulfill 
this idea in three steps: First, we identify the tokens that should be masked in the cloze task. Intuitively, such tokens need to indicate the answers to the original questions.
Then, we construct the prompt data by combining the masked tokens and our designed template. Finally, we feed the original QA training samples and the created prompt data into a pre-trained BART model~\cite{lewis2020bart} for fine-tuning.

\vspace{-10pt}
\subsection{Entity-aware Text Masking}
The first step to fulfill the cloze task is entity-aware text masking. The cloze task is often referred to as ``masked language modeling'' (MLM) in the literature~\cite{devlin2019bert}. Although MLM is widely used as a pre-training task in NLP, it is still less explored how to pick the masked token spans to achieve good performance in a specific downstream task. Indeed, prevailing PLMs like BERT~\cite{devlin2019bert} randomly mask a proportion of tokens in each sequence.
However, even though PLMs with randomly masking statistically can survive with a large-scale training corpus and the law of large numbers, few-shot QA tasks with only tens of short sequences could potentially receive only weak and noisy samples.
For the sake of our QA task, we propose to enable the model to infer the crucial parts of the reasoning procedure. Human reasoning is usually considered as hops between entities~\cite{cai2021deep}. A robust model should be able to recover important masked entities based on their context. To achieve this idea, we take the entity set of the Wikidata knowledge graph~\cite{vrandevcic2014wikidata} as the entity corpus. For each training sample, we extract all the text spans recognized as Wikidata entities.
As a result, the created cloze questions will be centered around meaningful entities rather than irrelevant tokens to the QA task, such as articles, pronouns, and stop-words.

\vspace{-10pt}

\subsection{Prompt-style Data Augmentation}

Based on the output of entity-aware text masking, we pursue the recent success of prompt-tuning to produce augmented data for the prompt-based \textsc{Gotta} model.
Specifically, we formally formulate the following template to integrate QA and cloze tasks, thereby generating few-shot QA input data $x^{ori}$ as:
\begin{gather*}
    x_{q}= \emph{Question}: \mathbf{q} \\
    x_{a}= \emph{Answer}: \langle {\rm mask} \rangle \\
    x_{c}= \emph{Context}: \mathbf{c} \\
    x^{ori}=\left[x_{q} \oplus x_{a} \oplus x_{c}\right]
\end{gather*}
The labels $y$ are formulated as follows:
\begin{gather*}
    y_{a}= \emph{Answer}: \mathbf{a}, \\
    y=\left[x_{q} \oplus y_{a} \oplus x_{c}\right],
\end{gather*}
where $\mathbf{q}$, $\mathbf{a}$ and $\mathbf{c}$ are texts of the question, answer text, and context, respectively; $\oplus$ denotes string concatenation.

For the augmented data, we fix the question text~$\mathbf{q^{\text{aug}}}$ as follows:
\begin{gather*}
    \mathbf{q^{\text{aug}}}= \emph{What is the masked entity?}
\end{gather*}
Note that in the augmented cloze data samples, we also mask the selected entity in $x_{c}$ to form the context text for the augmented data $x_{c}^{\emph{aug}}$ in addition to the mask token in $x_{a}$.
Figure~\ref{fig:aug_data_example} illustrates the details of an augmented data sample $(x^{\emph{aug}}, y^{\emph{aug}})$. Let $(X^{\emph{ori}}, Y^{\emph{ori}})$ and $(X^{\emph{aug}}, Y^{\emph{aug}})$ denote all the training samples of QA and cloze, respectively. Our complete training set $(X^{\emph{train}}, Y^{\emph{train}})$ is the union of $(X^{\emph{ori}}, Y^{\emph{ori}})$ and $(X^{\emph{aug}}, Y^{\emph{aug}})$.

\subsection{Generative Prompt-Tuning}

One of the most apparent advantages of aligning augmented and original data is the model's capability of seamlessly digesting both without a distinct loss.
In a nutshell, \textsc{Gotta} adopts an encoder-decoder model as:
\begin{equation}
    y^{\emph{pred}} = {\rm \emph{decoder}}_{\theta_{\emph{D}}}({\rm \emph{encoder}}_{\theta_{\emph{E}}}(x)),
\end{equation}
where $\theta_{\emph{E}}$ and $\theta_{\emph{D}}$ are learnable parameters; $x \in X^{train}$ can be either an original training sample or an augmented one.

Our training objective maximizes the log-likelihood of the text in the reference answer $y \in Y^{train}$. 
The loss functions with respect to the original samples and the augmented samples can be expressed as follows:
\begin{multline}
L^{\emph{ori}}(\theta)= \\
\sum_{\left(x, y\right) \in\left(X^{\emph{ori}}, Y^{\emph{ori}}\right)} \log \left(\prod_{i=1}^{n} P\left(y_{i} \mid y_{<i}, x; \theta\right)\right)
\end{multline}
and 
\begin{multline}
L^{aug}(\theta)= \\
\sum_{\left(x, y\right) \in\left(X^{\emph{aug}}, Y^{\emph{aug}}\right)} \log \left(\prod_{i=1}^{n} P\left(y_{i} \mid y_{<i}, x; \theta\right)\right),
\end{multline}
where $\theta = \{\theta_{D}, \theta_{E}\}$.

The overall loss function takes a weighted sum:
\begin{equation}
    L(\theta) = L^{\emph{ori}}(\theta) + \lambda L^{\emph{aug}}(\theta).
\end{equation}
Here, $\lambda >0$ is a hyperparameter that balances between the QA task and the prompted cloze task.

\section{Experiments}
\label{sec:exp}

In this section, we describe in detail how we set up our experiments, then we report the experimental results and discuss the results. We further provide some in-depth analysis of \textsc{Gotta}, through which we can better understand the model.

\subsection{Experimental Setup}
\noindent \textbf{Datasets.}
Following Splinter~\cite{ram2021few} and FewshotQA~\cite{chada2021fewshotqa}, we sample subsets from the MRQA 2019 shared task~\cite{fisch2019mrqa} for our few-shot experiments.
Specifically, MRQA contains eight widely used benchmark question answering datasets: SQuAD~\cite{rajpurkar2016squad}, NewsQA~\cite{trischler2017newsqa}, TriviaQA~\cite{joshi2017triviaqa}, SearchQA~\cite{dunn2017searchqa}, HotpotQA~\cite{yang2018hotpotqa}, Natural Questions~\cite{kwiatkowski2019natural}, BioASQ~\cite{tsatsaronis2015overview}, and TextbookQA~\cite{kembhavi2017you}. Following Splinter~\cite{ram2021few}, smaller training datasets are sampled in a logarithmic manner from the original full datasets, resulting in few-shot datasets with training example numbers 16, 32, 64, and 128.

\vspace{5pt}

\noindent \textbf{Comparative Baselines.} We evaluate the performance of \textsc{Gotta} against four competitive few-shot QA methods, including  \textbf{RoBERTa~}\cite{liu2019roberta}, \textbf{SpanBERT}~\cite{joshi2020spanbert}, \textbf{Splinter}~\cite{ram2021few}, and \textbf{ FewshotQA}~\cite{chada2021fewshotqa}.

\vspace{1mm}

\noindent \textbf{Implementation Details.}
We extract $24,863,792$ entities from Wikidata for entity candidate matching. When extracting the entities in the contexts of training samples, we use the Aho-Corasick algorithm\footnote{https://github.com/WojciechMula/pyahocorasick/} \cite{aho1975efficient} to conduct exact multi-pattern lexical matching.
For all the models, we use the same hyperparameters during training for a fair comparison. Specifically, the models are optimized by Adam \cite{kingma2014adam} with bias corrections. The learning rate is $2 \times 10^{-5}$ without learning rate scheduling. The training batch size is set to $2$. The maximum sequence length of sequence generation is $100$ for FewshotQA and \textsc{Gotta}. We train all compared models for $25$ epochs. The reported results are given by the best-performing checkpoint on the development sets. For \textsc{Gotta}, we perform a grid search for the loss weight $\lambda$ in the space $\{0.01, 0.05, 0.1, 0.5, 1.0, 10.0\}$. 
All the experiments are run on NVIDIA Tesla A100-SXM4 Tensor Core GPUs with 40GB memory.

\begin{table*}[t!]
\centering
\resizebox{.9\linewidth}{!}{
\begin{tabular}{ccccccccc}
\toprule
\textbf{\# examples} & \textbf{SQuAD} & \textbf{TriviaQA} & \textbf{NQ}   & \textbf{NewsQA} & \textbf{SearchQA} & \textbf{HotpotQA} & \textbf{BioASQ} & \textbf{TextbookQA} \\ \midrule
16         & 336   & 2,118     & 883  & 1,904   & 2,620     & 517      & 591    & 1,814       \\
32         & 711   & 4,287     & 1,422 & 2,801   & 5,452     & 1,005     & 1,205   & 3,934       \\
64         & 1,539  & 8,592     & 2,696 & 5,867   & 10,601    & 2,090     & 2,568   & 7,526       \\
128        & 3,052  & 17,301    & 4,989 & 11,469  & 21,113    & 4,128     & 5,226   & 15,504     \\ \bottomrule
\end{tabular}}
\caption{Number of augmented training examples per dataset. We construct one training example for each entity extracted from the passages and form the cloze task.}
\vspace{-10pt}
\label{tab:aug_data}
\end{table*}

\vspace{1mm}

\noindent \textbf{Evaluation Metrics.}
Following previous studies~\cite{ram2021few,chada2021fewshotqa}, we use the F1 score as our evaluation metric. Specifically, for each sample in the test set, the predicted span and the ground truth answer are treated as bags of words, and F1 scores are applied to compute the overlap between these two sets. If there are multiple ground-truth answers to a particular question, we take the maximum of the corresponding F1 scores.

\subsection{Performance Comparison}
Table \ref{tab:overall_perf} shows the few-shot QA performance of compared models across all the benchmarks when 16, 32, 64, and 128 training examples are given. 
For both \textbf{FewshotQA} and \textbf{\textsc{Gotta}}, we use BART-large as the backbone PLM. We also report their performance when BART-base is applied as the PLM, in which case the models are denoted as \textbf{FewshotQA-base} and \textbf{\textsc{Gotta}-base}, respectively.
We repeat the same experiment $5$ times using different random seeds and report the mean and standard deviation of the results for each method.
Furthermore, we include the relative performance gain of \textsc{Gotta} over the second-best method, i.e., FewshotQA. Overall, \textsc{Gotta} outperforms all the compared methods by a decent margin in most cases. Even beyond that, we observe a lower variance in results produced by \textsc{Gotta} over FewshotQA in most cases (24 out of 32), especially when fewer training examples are available (14 out of 16).

Next, let us take a closer look at specific datasets. On SQuAD and HotpotQA, \textsc{Gotta} consistently achieves higher F1 with lower variance. On TriviaQA, NewsQA, SearchQA, and TextbookQA, we observe relatively more significant performance gains over the best baseline. We conjecture that it is because the number of augmented data samples on these datasets is larger than that on other datasets. Therefore, signals from the cloze task are sufficient to impact the main QA task positively.

\begin{table*}[t!]
\centering
\resizebox{.9\linewidth}{!}{
\begin{tabular}{lcccccccc}
\toprule
\multicolumn{1}{c}{\textbf{Model}} & \multicolumn{1}{c}{\textbf{SQuAD}}    & \multicolumn{1}{c}{\textbf{TriviaQA}} & \multicolumn{1}{c}{\textbf{NQ}}       & \multicolumn{1}{c}{\textbf{NewsQA}}   & \multicolumn{1}{c}{\textbf{SearchQA}} & \multicolumn{1}{c}{\textbf{HotpotQA}} & \multicolumn{1}{c}{\textbf{BioASQ}}   & \multicolumn{1}{c}{\textbf{TextbookQA}}                         \\ \toprule
\multicolumn{9}{l}{16 Examples}           \\ \bottomrule
RoBERTa                         & 7.7±4.3  & 7.5±4.4  & 17.3±3.3 & 1.4±0.8  & 6.9±2.7  & 10.5±2.5 & 16.7±7.1 & 3.3±2.1  \\
SpanBERT                        & 18.2±6.7 & 11.6±2.1 & 19.6±3.0 & 7.6±4.1  & 13.3±6.0 & 12.5±5.5 & 15.9±4.4 & 7.5±2.9  \\
Splinter                        & 54.6±6.4 & 18.9±4.1 & 27.4±4.6 & 20.8±2.7 & 26.3±3.9 & 24.0±5.0 & 28.2±4.9 & 19.4±4.6 \\ \midrule
FewshotQA-base                  & 55.3±2.7 & 39.6±6.2 & 46.9±1.4 & 36.5±2.6 & 40.8±4.4 & 43.7±2.4 & 52.1±1.6 & 16.7±2.2 \\
FewshotQA                       & 72.5±3.7 & 47.1±7.6 & 57.3±3.2 & 44.9±4.5 & 54.3±5.9 & 59.7±2.2 & 62.7±4.4 & 33.1±3.2 \\ \midrule
\textsc{Gotta}-base             & 57.8±2.6 & 40.8±5.6 & 47.1±1.1 & 36.2±1.6 & 41.8±5.4 & 45.9±1.7 & 55.2±2.5 & 20.5±1.9 \\
\textsc{Gotta}                  & \textbf{74.6±1.9} & \textbf{63.3±8.0} & \textbf{58.9±1.9} & \textbf{47.3±2.5} & \textbf{56.8±3.9} & \textbf{59.8±2.1} & \textbf{66.1±3.1} & \textbf{38.5±5.3} \\  \toprule
Improvement\%                     & 2.9      & 34.3     & 2.8      & 5.3      & 4.5      & 0.1      & 5.4      & 16.1     \\ \toprule
32 Examples                     &          &          &          &          &          &          &          &          \\ \bottomrule
RoBERTa                         & 18.2±5.1 & 10.5±1.8 & 22.9±0.7 & 3.2±1.7  & 13.5±1.8 & 10.4±1.9 & 23.3±6.6 & 4.3±0.9  \\
SpanBERT                        & 25.8±7.7 & 15.1±6.4 & 25.1±1.6 & 7.2±4.6  & 14.6±8.5 & 13.2±3.5 & 25.1±3.3 & 7.6±2.3  \\
Splinter                        & 59.2±2.1 & 28.9±3.1 & 33.6±2.4 & 27.5±3.2 & 34.8±1.8 & 34.7±3.9 & 36.5±3.2 & 27.6±4.3 \\ \midrule
FewshotQA-base                  & 59.5±2.2 & 50.3±3.1 & 48.1±2.1 & 40.7±2.3 & 49.4±3.2 & 48.2±1.7 & 56.7±2.2 & 24.1±4.2 \\
FewshotQA                       & 73.8±2.2 & 56.7±5.9 & \textbf{60.6±2.4} & 50.0±2.8 & 61.4±3.6 & 61.6±1.5 & 66.9±4.7 & 41.7±4.2 \\ \midrule
\textsc{Gotta}-base             & 62.7±1.8 & 47.7±4.5 & 49.6±1.3 & 41.4±2.4 & 49.8±2.5 & 49.6±1.3 & 57.6±3.0 & 28.1±1.9 \\
\textsc{Gotta} & \textbf{76.0±2.0} & \textbf{61.9±4.8} & 59.8±2.4 & \textbf{51.2±1.5} & \textbf{63.1±3.1} & \textbf{62.7±1.2} & \textbf{69.5±1.0} & \textbf{46.3±3.7} \\  \toprule
Improvement\%                     & 3.0      & 9.1      & -1.4     & 2.4      & 2.8      & 1.7      & 3.8      & 11.1     \\ \toprule
64 Examples                     &          &          &          &          &          &          &          &          \\ \bottomrule
RoBERTa                         & 28.4±1.7 & 12.5±1.4 & 24.2±1.0 & 4.6±2.8  & 19.8±2.4 & 15.0±3.9 & 34.0±1.8 & 5.4±1.1  \\
SpanBERT                        & 45.8±3.3 & 15.9±6.4 & 29.7±1.5 & 12.5±4.3 & 18.0±4.6 & 23.3±1.1 & 35.3±3.1 & 13.0±6.9 \\
Splinter                        & 65.2±1.4 & 35.5±3.7 & 38.2±2.3 & 37.4±1.2 & 39.8±3.6 & 45.4±2.3 & 49.5±3.6 & 35.9±3.1 \\ \midrule
FewshotQA-base                  & 66.5±1.1 & 52.3±2.8 & 51.5±1.6 & 43.5±2.0 & 54.9±2.0 & 50.7±1.6 & 64.3±2.3 & 31.7±2.8 \\
FewshotQA                       & 77.9±2.1 & 57.9±4.4 & 60.9±2.5 & 53.7±1.1 & 65.4±2.4 & 63.1±2.2 & 73.2±3.1 & 44.8±1.8 \\ \midrule
\textsc{Gotta}-base             & 67.7±0.9 & 50.6±4.0 & 51.5±1.3 & 45.7±1.6 & 54.6±3.1 & 52.0±0.8 & 64.9±2.6 & 35.5±3.5 \\
\textsc{Gotta} & \textbf{78.9±0.5} & \textbf{59.6±1.9} & \textbf{63.6±1.0} & \textbf{54.3±3.0} & \textbf{66.3±2.5} & \textbf{64.3±1.7} & \textbf{73.2±1.5} & \textbf{51.2±2.8} \\  \midrule
Improvement\%                     & 1.3      & 3.0      & 4.4      & 1.1      & 1.4      & 1.9      & 0.0     & 14.3     \\ \toprule
128 Examples                    &          &          &          &          &          &          &          &          \\ \midrule
RoBERTa                         & 43.0±7.1 & 19.1±2.9 & 30.1±1.9 & 16.7±3.8 & 27.8±2.5 & 27.3±3.9 & 46.1±1.4 & 8.2±1.1  \\
SpanBERT                        & 55.8±3.7 & 26.3±2.1 & 36.0±1.9 & 29.5±7.3 & 26.3±4.3 & 36.6±3.4 & 52.2±3.2 & 20.9±5.1 \\
Splinter                        & 72.7±1.0 & 44.7±3.9 & 46.3±0.8 & 43.5±1.3 & 47.2±3.5 & 54.7±1.4 & 63.2±4.1 & 42.6±2.5 \\ \midrule
FewshotQA-base                  & 70.8±0.7 & 45.9±2.1 & 53.6±1.1 & 48.4±1.8 & 58.7±0.9 & 56.3±0.9 & 73.8±1.0 & 37.7±1.1 \\
FewshotQA                       & 78.8±2.7 & 55.2±1.8 & 63.3±1.6 & 56.8±1.1 & 67.0±1.8 & 64.9±1.8 & 77.2±1.5 & 46.2±5.9 \\ \midrule
\textsc{Gotta}-base             & 71.3±1.3 & 52.8±2.0 & 54.2±0.7 & 49.8±1.6 & 60.2±1.6 & 56.3±1.4 & 73.1±1.9 & 40.3±3.2 \\
\textsc{Gotta} & \textbf{80.8±1.7} & \textbf{60.0±3.6} & \textbf{64.9±1.2} & \textbf{57.4±1.2} & \textbf{69.8±1.5} & \textbf{66.7±1.8} & \textbf{78.6±2.1} & \textbf{53.3±1.7} \\ \midrule
Improvement\%                     & 2.6      & 8.8      & 2.5      & 1.1      & 4.3      & 2.9      & 1.8      & 15.3     \\ \bottomrule
\end{tabular}}
\caption{Overall performance in F1 scores across all datasets when the numbers of training examples are 16, 32, 64, and 128. NQ stands for Natural Questions. Improvement\% marks the relative performance improvements of \textsc{Gotta} compared to the best baselines. RoBERTa, SpanBERT, and Splinter have 110M parameters. FewshotQA-base and \textsc{Gotta}-base have 130M parameters. Both FewshotQA and \textsc{Gotta} have parameters of size 406M. The average improvements of \textsc{Gotta} over FewshotQA are significant on all eight datasets in a paired t-test (p-value $<$ 0.05).}
\vspace{-6pt}
\label{tab:overall_perf}
\end{table*}

\begin{table*}[t!]
\centering
\resizebox{.9\linewidth}{!}{
\begin{tabular}{lcccccccc}\toprule
\textbf{Model} &\textbf{SQuAD} &\textbf{TriviaQA} &\textbf{NQ} &\textbf{NewsQA} &\textbf{SearchQA} &\textbf{HotpotQA} &\textbf{BioASQ} &\textbf{TextbookQA} \\ \midrule
\multicolumn{9}{l}{16 Examples} \\ \midrule
\textsc{Gotta} &\textbf{74.6±1.9} &\textbf{63.3±8.0} &\textbf{58.9±1.9} &\textbf{47.3±2.5} &\textbf{56.8±3.9} &59.8±2.1 &66.1±3.1 &38.5±5.3 \\
\textsc{Gotta}-random &72.1±2.6 &53.2±8.4 &56.2±4.1 &46.7±2.2 &54.8±6.1 &\textbf{61.2±1.0} &61.9±2.3 &38.4±2.7 \\
\textsc{Gotta}-MTL &71.0±1.9 &49.4±7.7 &57.8±2.6 &45.1±3.5 &56.0±5.1 &58.4±2.3 &62.9±4.5 &37.4±3.4 \\
\textsc{Gotta}-what &69.8±2.9 &52.0±7.3 &57.9±3.3 &46.8±1.8 &54.9±4.4 &60.1±1.0 &\textbf{66.2±3.3} &\textbf{38.8±2.3} \\ \midrule
\multicolumn{9}{l}{32 Examples} \\ \midrule
\textsc{Gotta} &\textbf{76.0±2.0} &\textbf{61.9±4.8} &59.8±2.4 &51.2±1.5 &\textbf{63.1±3.1} &62.7±1.2 &69.5±1.0 &\textbf{46.3±3.7} \\
\textsc{Gotta}-random &75.9±2.1 &54.7±5.4 &59.3±1.7 &\textbf{51.5±2.2} &62.8±2.3 &\textbf{63.3±1.6} &67.5±3.8 &42.6±4.9 \\
\textsc{Gotta}-MTL &70.9±2.4 &55.5±5.8 &60.0±1.4 &48.7±3.2 &60.8±1.7 &61.4±1.2 &66.7±1.9 &41.5±3.6 \\
\textsc{Gotta}-what &74.7±1.1 &54.6±5.6 &\textbf{60.4±2.2} &50.0±1.2 &62.4±2.9 &60.2±1.7 &\textbf{70.7±1.3} &40.4±4.1 \\ \midrule
\multicolumn{9}{l}{64 Examples} \\ \midrule
\textsc{Gotta} &78.9±0.5 &\textbf{59.6±1.9} &\textbf{63.6±1.0} &\textbf{54.3±3.0} &66.3±2.5 &\textbf{64.3±1.7} &\textbf{73.2±1.5} &\textbf{51.2±2.8} \\
\textsc{Gotta}-random &\textbf{79.3±1.3} &57.9±3.4 &62.2±1.6 &53.0±3.0 &66.1±3.4 &63.8±1.7 &72.8±1.7 &51.1±3.3 \\
\textsc{Gotta}-MTL &73.9±2.7 &54.5±5.0 &60.7±1.1 &52.6±1.1 &65.7±2.3 &63.3±1.7 &71.6±2.6 &45.1±3.3 \\
\textsc{Gotta}-what &78.7±1.2 &59.2±2.5 &62.7±0.9 &54.2±1.6 &\textbf{67.2±1.3} &64.0±1.0 &70.9±3.4 &48.0±1.9 \\ \midrule
\multicolumn{9}{l}{128 Examples} \\ \midrule
\textsc{Gotta} &80.8±1.7 &\textbf{60.0±3.6} &\textbf{64.9±1.2} &57.4±1.2 &\textbf{69.8±1.5} &\textbf{66.7±1.8} &\textbf{78.6±2.1} &\textbf{53.3±1.7} \\
\textsc{Gotta}-random &79.9±1.0 &58.6±4.0 &64.3±0.9 &57.2±0.9 &69.8±1.5 &66.0±0.7 &78.1±2.1 &52.5±4.1 \\
\textsc{Gotta}-MTL &77.2±1.9 &54.1±1.7 &62.4±1.0 &53.1±2.1 &65.9±1.9 &64.1±1.9 &76.5±1.3 &47.6±2.1 \\
\textsc{Gotta}-what &\textbf{80.9±1.4} &57.8±4.0 &64.5±0.6 &\textbf{57.6±0.6} &67.5±0.9 &64.7±1.5 &77.7±1.9 &52.4±2.3 \\
\bottomrule
\end{tabular}
}
\vspace{-5pt}
\caption{Performance of different model variations across all datasets in F1 scores. We also conduct significant tests for \textsc{Gotta}-random. However, \textsc{Gotta}-random does not significantly outperform FewshotQA (p-value $\gg$ 0.05).}
\vspace{-5pt}
\label{tab:ablation_perf}
\end{table*}

\subsection{Analysis and Discussions}

We further provide more in-depth studies to look into which steps and parts contribute the most to \textsc{Gotta}'s performance.
Looking back on the design of our model, three key modules are proposed, namely entity masking, prompt data construction, and prompt loss design.
Besides, data augmentation also plays a vital role in \textsc{Gotta}.


\subsubsection{Entity Masking}

We start from the entity masking module. To check whether entity masking benefits the overall performance, we create a variation of \textsc{Gotta} called \textsc{Gotta}-random. In \textsc{Gotta}-random, we remove the entity masking module and randomly mask text spans instead of entities that appear in the Wikidata entity set. As shown in Table~\ref{tab:ablation_perf} comparing between \textsc{Gotta} and \textsc{Gotta}-random, we find that: (1) Randomly masking usually yields a higher variance. Although the cloze task can still be fulfilled by randomly selecting phrases, it destabilized the overall QA performance. (2) The full model outperforms the random model in most cases, which validates our hypothesis that masking entities in the context are crucial for selecting the subjects of the cloze examples, thus improving the QA task.

\subsubsection{Prompt-tuning vs. Multi-task Learning}
Prompt data construction is the second key step proposed in our \textsc{Gotta} framework.
As an analysis, we compare prompt tuning with multi-task learning, which can be the other intuitive approach to jointly learn the QA and cloze tasks.
Specifically, we denote \textsc{Gotta} with multi-tasking learning as \textsc{Gotta}-MTL.

From Table~\ref{tab:ablation_perf}, we observe that (1) \textsc{Gotta}-MTL has apparently worse performance than \textsc{Gotta}, which validates our claim that formulating the cloze task in the same format of QA is essential. (2) \textsc{Gotta}-MTL is defeated by \textsc{Gotta}-what in most cases, meaning that the contribution of prompt is larger than that of entity masking or question text. That being said, aligning the format of QA, cloze along with that of the pre-training task contributes the most to the overall performance.

\vspace{6pt}

\subsubsection{Question Templates}
Now that we have shown that it is necessary to formulate the cloze task as prompt-tuning, a natural question is: \emph{Does the question text have an impact on the prompt-tuning performance?} To answer this, we construct another model \textsc{Gotta}-what to study the effect of question text on the performance. The mere distinct between \textsc{Gotta}-what and the original model is the question text of the augmented data. Formally, we change the original question
\begin{gather*}
    \mathbf{q}= \emph{What is the masked entity?}
\end{gather*}
to the question
\begin{gather*}
    \mathbf{q}= \emph{What?}
\end{gather*}
Comparing the performance of \textsc{Gotta}-what with that of \textsc{Gotta} in Table~\ref{tab:ablation_perf}, we observe that the two are comparable. On TriviaQA, \textsc{Gotta} slightly outperforms \textsc{Gotta}-what consistently while things are otherwise on any other dataset, with the two going back and forth.

\begin{figure*}[t!]
    \centering
    \includegraphics[width=.9\linewidth]{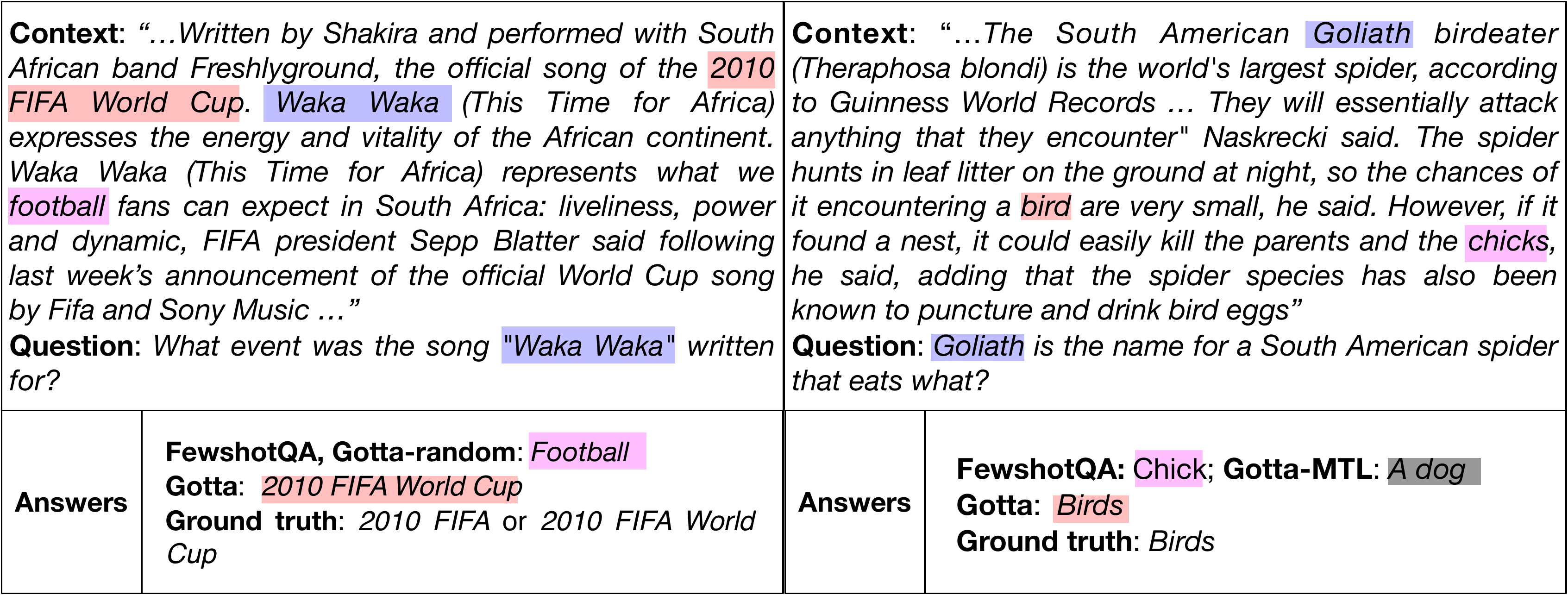}
    \vspace{-10pt}
    \caption{Answers generated by different models for two test cases from TriviaQA. We match the color of the generated answers with their occurrences in the text if they are in the text. In both cases, \textsc{Gotta} successfully generates the correct answer, whereas baselines without entity masking can not accurately recover the entity-level details. }
    \label{fig:case_study}
    \vspace{-10pt}
\end{figure*}


\vspace{-6pt}

\subsubsection{Case Study}
We further take a look at two concrete test cases. Figure~\ref{fig:case_study} illustrates two examples sampled from the test set of TriviaQA. As we can see, in the left case, both FewshotQA and \textsc{Gotta}-random generate the incorrect answer \emph{Football}. While this generated answer has highly relevant semantics to the correct answer \emph{2010 FIFA World Cup}, that answer is still not detailed enough. From this observation, we validate our claim that compared with FewshotQA and \textsc{Gotta}-random without an entity masking module, the full model of \textsc{Gotta} can generate the answer text in detail from the entity level. In the right case, \textsc{Gotta} generates the correct answer. In contrast, although \textsc{Gotta}-MTL generates the answer \emph{A dog} that a spider could eat, it is still a wrong answer and does not even appear in the context.
This difference perfectly demonstrates that prompt-tuning is beneficial to building connections between entities in the same context. Although FewshotQA returns an answer within the context, the answer is too trivial to answer the question.
These two cases provide evidence to validate that entity-aware masking and prompt-style data augmentation in our proposed \textsc{Gotta} are both essential to acquiring the capability of deeply understanding the complicated semantics in questions and contexts.

\vspace{-8pt}

\subsubsection{Effect of Augmented Data}

\begin{figure}[htbp]
    \centering
    \includegraphics[width=.9\linewidth]{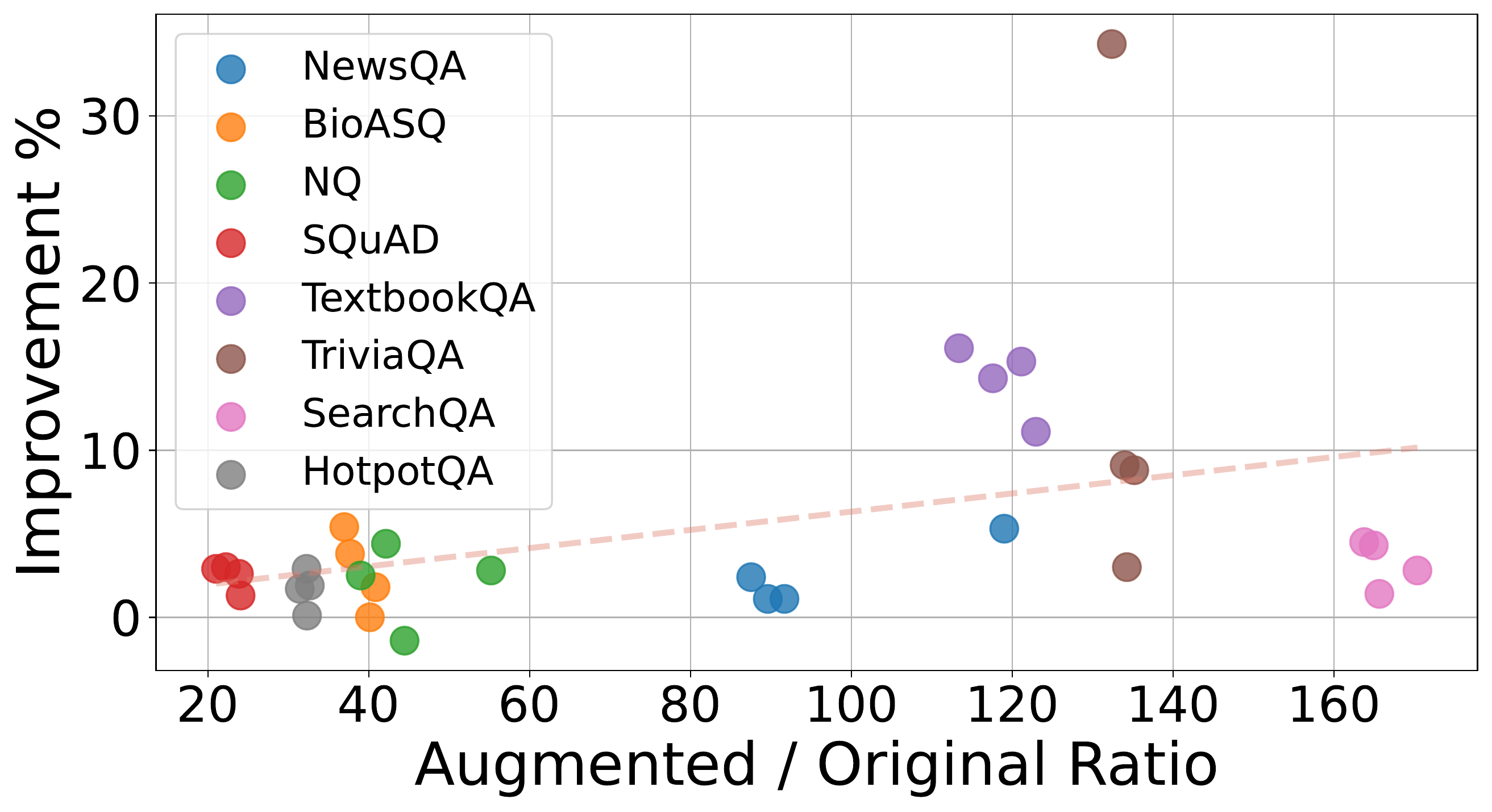}
    \vspace{-5pt}
    \caption{Relative performance improvement w.r.t. the number of augmented data generated per sample. There are in total 32 data points corresponding to each setting on each dataset in Table~\ref{tab:overall_perf}.}
    \label{fig:gain_analysis}
    \vspace{-15pt}
\end{figure}

As shown in Figure~\ref{fig:gain_analysis}, we proceed to study the actual effect of augmented data on the overall performance by investigating the relationship between \textbf{average augmented example per training example} and \textbf{relative performance improvement}. With the growth of average augmented data per training example, the performance gain is generally larger. Recall that we construct augmented data by raising questions on the entities detected in the context of training examples. When there are more entities in the context, \textsc{Gotta} can learn more about the semantics of the entities and potentially the relations in between, thus having a deeper understanding of the context, thereby further strengthening the QA performance. However, we do observe there is not much gain on SearchQA. Our conjecture is that the contexts of SearchQA are usually very long, so it is rather hard to match the most critical entities. In the extreme case, the entity masking degenerates to MLM, omitting the role of the entities.

\vspace{-5pt}

\section{Conclusion and Future Work}
\label{sec:conclusion}
In this work, we propose to incorporate the cloze task to improve neural machine question answering with a few training examples. The key idea is to identify and mask the informative entities in the passage and make the model predict them correctly. Through empirical experimental studies on various QA benchmarks and different few-shot settings, we show that the cloze task indeed benefits the QA task due to its commonalities. We find different ways of incorporating the cloze task improve the QA task while prompt-tuning brings the most. Looking forward, it is of interest to explore QA-dedicated pre-training and ways of pipelining pre-training and prompt-tuning for downstream few-shot QA needs.

\vspace{-10pt}

\section*{Acknowledgements}
This paper was partially supported by NSF 1829071, 2106859, 2200274, Cisco and NEC. We thank Ruihan Wu for the useful discussions on the implementation of the framework. 

\vspace{-10pt}
\bibliographystyle{siam}
\bibliography{sdm}

\appendix

\section{Comparative Baselines}
\label{appendix:baselines}

In this section, we present the details of four competitive baseline methods in our experiments.

\begin{itemize}[leftmargin=*]
\item \textbf{RoBERTa \cite{liu2019roberta}} is a robustly optimized BERT-based PLM. It improves BERT by techniques such as training the model for a longer time, with larger batches, and getting rid of the next sentence prediction task. It is known to demonstrate substantially better performance on a variety of natural language understanding tasks over BERT, including question answering.

\item \textbf{SpanBERT \cite{joshi2020spanbert}} is another variant of BERT that emphasizes on the encoding of spans instead of tokens. It is pre-trained on two tasks: (1) masked language modeling, which is the same as BERT, and (2) span boundary prediction, which pulls the representations of the span boundary into a direction where the entire content of the masked span can be predicted correctly. SpanBERT reaches substantially better performance on span selection tasks in particular.

\item \textbf{Splinter \cite{ram2021few}} is a pre-training framework dedicated to the extractive QA task based on SpanBERT. It is pre-trained by the recurring span selection task, which masks all but one instance of each recurring span and asks the model to select the correct span for each masked position.

\item \textbf{FewshotQA \cite{chada2021fewshotqa}} is the first QA-dedicated fine-tuning framework that leverages pre-trained encoder-decoder models such as BART~\cite{lewis2020bart} and T5~\cite{raffel2020exploring}. In FewshotQA, the input is constructed as a concatenation of the question, a mask token as the placeholder for the answer span, and a context. Given this input, the model is fine-tuned using the same objective as its pre-training objective.
\end{itemize}

\begin{figure}[t!]
    \centering
    \includegraphics[width=.65\linewidth]{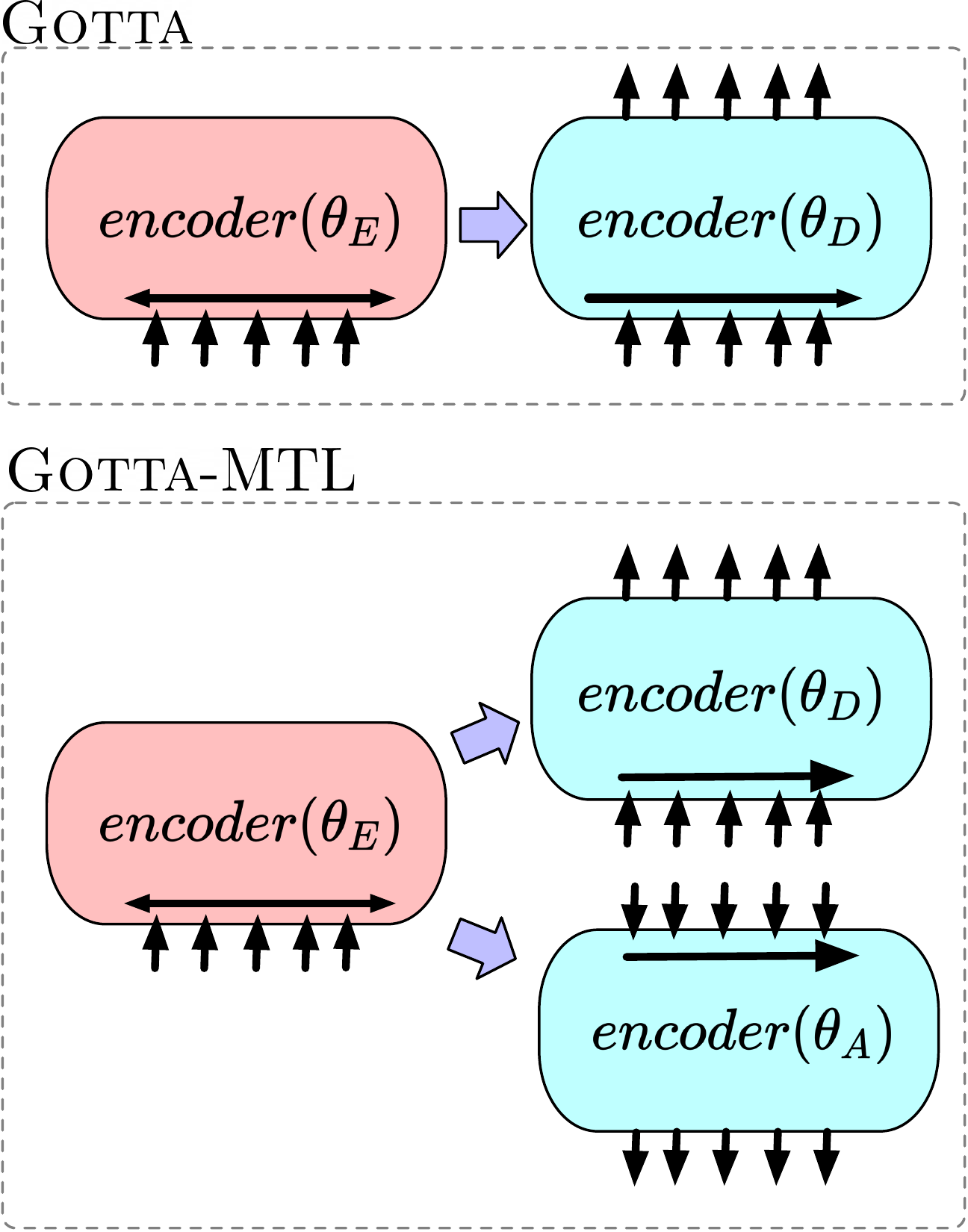}
    \caption{Head-to-head comparison between the architectures of \textsc{Gotta} and \textsc{Gotta}-MTL.}
    \label{fig:vs}
\end{figure}

\section{Multi-task Learning Variant}
\label{appendix:mtl}
Apart from prompt-based data augmentation, we also come up with a multi-task learning (MTL) based variant of \textsc{Gotta}. The reason why we implement this variant is that MTL has been proved capable of exploiting the commonalities and differences across tasks. We expect both QA and cloze can be improved when MTL is applied. We refer to the MTL variant as \textsc{Gotta}-MTL.

We demonstrate the model architectures of \textsc{Gotta} and \textsc{Gotta}-MTL in Figure~\ref{fig:vs}. 
The major difference between these two architectures is on the decoder side. 
To be specific, \textsc{Gotta}-MTL implements MTL by separating the decoders for QA and cloze. In \textsc{Gotta}-MTL, while the original QA training samples and the augmented cloze training samples go through the same encoder, each of the two tasks has its exclusive decoder. Formally, the forward process of \textsc{Gotta}-MTL is as follows:
\begin{equation}
\begin{split}
    y^{ori}_{pred} = decoder_{\theta_{\emph{D}}}(encoder_{\theta_{\emph{E}}}(x_{ori})), \\
    y^{aug}_{pred} = decoder_{\theta_{\emph{A}}}(encoder_{\theta_{\emph{E}}}(x_{\emph{aug}})), \\
    L(\theta) = L^{ori}(\theta_{\emph{E}}, \theta_{\emph{D}}) + \lambda L^{aug}(\theta_{\emph{E}}, \theta_{\emph{A}}),
\end{split}
\end{equation} 
where $\theta = \{\theta_{\emph{E}}, \theta_{\emph{D}}, \theta_{\emph{A}}\}$.

\end{document}